\documentclass[conference]{IEEEtran}
\usepackage{url}
\usepackage{cite}
\usepackage{amsmath,amssymb,amsfonts}
\usepackage{algorithmic}
\usepackage{graphicx}
\usepackage{textcomp}
\usepackage{epsfig}
\usepackage{epsf}
\usepackage{bmpsize}
\usepackage{boxedminipage}
\usepackage[table]{xcolor}
\usepackage{ulem}

\usepackage{placeins}
\usepackage{listings}
\usepackage{lipsum}
\definecolor{codegreen}{rgb}{0,0.6,0}
\definecolor{codegray}{rgb}{0.5,0.5,0.5}
\definecolor{codepurple}{rgb}{0.58,0,0.82}
\definecolor{backcolour}{rgb}{0.95,0.95,0.92}

\lstdefinestyle{mystyle}{
  backgroundcolor=\color{white}, 
  keywordstyle=\color{magenta},
  numberstyle=\tiny\color{codegray},
  stringstyle=\color{codepurple},
  basicstyle=\ttfamily\footnotesize,
  breakatwhitespace=false,         
  breaklines=true,                 
  captionpos=b,                    
  keepspaces=true,                 
  numbers=left,                    
  numbersep=5pt,                  
  showspaces=false,                
  showstringspaces=false,
  showtabs=false,                  
  tabsize=2
}

\usepackage{listings}

\begin{document}
%


\title{mt5se: An Open Source Framework for Building Autonomous Trading Robots}

\author{\IEEEauthorblockN{Paulo Andr\'e Lima de Castro}
\IEEEauthorblockA{\textit{Autonomous Computational Systems Lab - LABSCA}\\
\textit{Aeronautics Institute of Technology (ITA - Instituto Tecnológico de Aeronáutica)} \\
S\~ao Jos\'e dos Campos-SP, Brazil \\
fidel.est@gmail.com,pauloac@ita.br}
}

\maketitle

\begin{abstract}
Autonomous trading robots have been studied in artificial
intelligence area for quite some time. Many AI techniques have been tested for building autonomous agents able to trade financial assets. These initiatives include traditional neural networks, fuzzy logic, reinforcement learning but also more recent approaches like deep neural networks and deep reinforcement learning. Many developers claim to be successful in creating robots with great performance when simulating execution with historical price series, so called backtesting. However, when these robots are used in real markets frequently they present poor performance in terms of risks and return. In this paper, we propose an open source framework (mt5se) that helps the development, backtesting, live testing and real operation of autonomous traders. We built and tested several traders using mt5se. The results indicate that it may help the development of better traders. Furthermore, we discuss the simple architecture that is used in many studies and propose an alternative multiagent architecture. Such architecture separates two main concerns for portfolio manager (PM) : price prediction and capital allocation. More than achieve a high accuracy, a PM should increase profits when it is right and reduce loss when it is wrong. Furthermore, price prediction is highly dependent of asset’s nature and history, while capital allocation is dependent only on analyst’s prediction performance and assets’ correlation. Finally, we discuss some promising technologies in the area. 
\end{abstract}


\section{Introduction}
\label{sec:intro}

An autonomous trader must be able of selecting, buying and selling financial assets in order to fulfill it’s investor requirements about risk and return within the investor’s horizon of
investment. This period may range from very some seconds or fractions of seconds (high frequency trading~\cite{Olympia:16}, to longer periods as several days or even years. It is an autonomous agent that perceives and acts in a very particular environment:
financial markets, which brings several complex features that may make it very challenging for autonomous agents. It is important to note that there are also ethical and legal implications that should be taken in consideration in the process of
building automated trading robots. We recommend Wellman and Rajan’s paper~\cite{Wellman:17} about ethical issues in Autonomous
Trading Agents.

In this paper, we propose  an \textbf{open source framework} framework for building and testing trading agents, called mt5se. We present some examples of trading agents developed using such architecture and the respective results. This framework allows the development of autonomous traders on python language, and the execution of \textbf{real-time simulation}. This kind of simulation using real-time data feed from real markets is very important to identify possible overfitting in financial machine learning models, as discussed in section~\ref{sec:overfitting}. The framework provides access to several stock exchanges (NYSE,Nasdaq, London SE, Tokyo SE, Brazil's B3 and others) through MetaTrader 5 platform~\cite{Metatrader:18}. Besides stock exchanges, it also allows trading with Foreign Exchanges pairs including cryptocurrencies like Bitcoin and Ethereum, through the so called Contract For Differences instrument~\cite{cfd:21}. Furthermore, it is possible to use the same autonomous trader with minor changes in real operations. It is possible to create Trading Robots using Neural networks, Random Forests, Support Vector Machines, Genetic Algorithms, Bayesian Networks, Reinforcement Learning, Deep Learning and other techniques using the wide range of available libraries in python. The mt5se framework is an evolution of a previous framework called mt5b3~\cite{mt5b3:20}, which is now deprecated.

\subsection{Organization of the Text}
{We assume the reader is already familiar with Finance Theory, including Modern Portfolio Theory~\cite{Markowitz:52}, Efficient Market Hypothesis~\cite{Fama:70}, Capital Asset Pricing Model (CAPM) and Market micro-structure. If that is not the case, a very short introduction to these concepts are available in section 2 of~\cite{Castro:20}}. The remainder of this paper is organized as follows: section~\ref{sec:modelling} discuss the main aspects of modelling autonomous analysts, traders and portfolio managers, including  some unique characteristics of  financial markets that make them hard challenges for artificial intelligence based agents. We present and analyze briefly some available frameworks for building autonomous traders and compare them with our proposed framework, called mt5se, which we present minutely	
 in section~\ref{sec:mt5se}. We also provide several examples of autonomous traders and portfolio managers and their source code in section~\ref{sec:results}.  We discuss some open problems in the development of high performance autonomous traders and we also present some AI related technologies that may contribute to the finance field in section~\ref{sec:probs}. Finally, we conclude the paper with some possible extensions and suggestion for future work in section~\ref{sec:conc}.

\section{Autonomous Analysts, Traders and Portfolio Managers}
\label{sec:modelling}

The spectrum of used AI techniques in finance field is wide and it includes since reinforcement Learning~\cite{Pereira:19,Stone:04}, multiagent systems~\cite{Castro:13,Wooldridge:18} complex networks~\cite{Zhao:18}, genetic algorithms~\cite{Stone:06}, random forests~\cite{Abdenur:19} to more recent approaches like deep reinforcement learning~\cite{Shoukry:21}. Regardless of the picked AI technology, there are some aspects that are always present and we observe that quite often AI researchers disregards significant aspects of finance theory, as for instance: the correlation among assets, different investor profiles and risk control. Dealing with such issues, it is fundamental to build better autonomous agents for finance, however it is also really challenging.

\subsection{Types of Financial Data}
\label{sec:types}

\begin{figure}[htb]
	\centering
		\includegraphics[width=0.90\columnwidth]{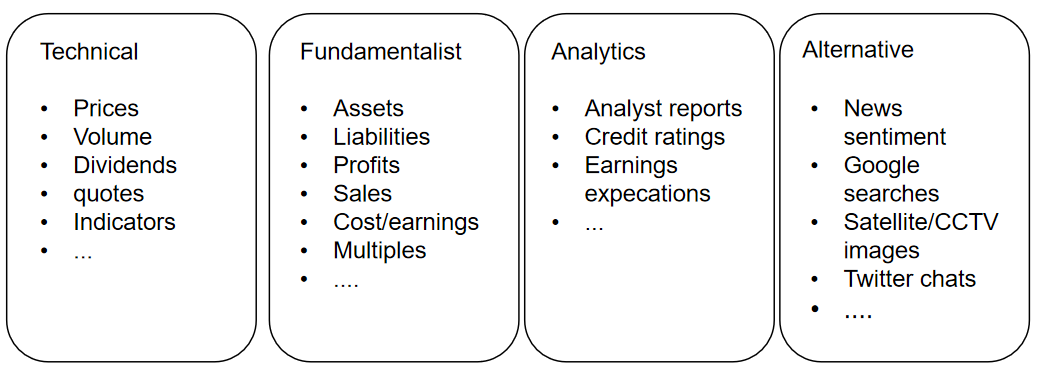}
	\caption{Types of Financial Data}
		\label{fig:types}
\end{figure}

The data that should be considered by an autonomous analyst is far more diverse and complex than historical prices or volumes. It can be splitted in four different types: technical, fundamentalist, analytical and alternative data~\cite{Prado:18}. Figure~\ref{fig:types} presents some examples for each type.

\textbf{Fundamental data} encompasses information that can be found in regulatory filings.  It is mostly accounting data, reported quarterly or yearly. A particular aspect of this data is that it is reported with some temporal lapse. You must confirm exactly when each data point was released, so that your analysis uses that information only after it was publicly available. A common error is to assume that this data was published at the end of the reporting period, which is almost never the case. Another problem is the use of back-filled or reinstated data. Back-filling means that missing data is assigned some value, even if those values were unknown at that time. A “reinstated value” is a corrected value that amends an incorrect initial release. The problem is, the corrected values were not known on that first release date. The use of such data gives a unrealistic advantage to models, that will not be present in real operation. In another words, we could say it also leads to overfitting. 
Nevertheless, it is common to find studies that use the final released value and assign it to the time of the first release, or even to the last day in the reporting period.

\textbf{Technical or Market} data includes all trading activity that takes place in an exchange, including open, high, low and closes prices and their historical series. It is common to use the term bar to refer to open, high, low and close prices in a given time frame (1 minute, 1 hour or 1 day ). Your Market data provider may also allow access to the order books or even to semi-structured information, like FIX messages. In this case, you may try to recognize patterns that identify some traders. For instance, human GUI traders often trade in round lots, and you can use this fact to estimate what percentage of the volume is coming from them at a given point in time.

\textbf{Analytics} data is usually provided by human experts. It is a kind of derivative data, based on an original source, which could be fundamental, market, alternative, or even a collection of other analytics. It is not available from an original source, and it was processed in a particular way (hopefully not biased). Independent analysts, investment banks and research firms sell valuable information that results from deep analyses of companies’ business models, activities, competition and so on. 

\textbf{Alternative data} may refer to satellite image or video feeds include monitoring of tankers, tunnel traffic activity, or parking lot occupancy.  Before an Oil company reports increased earnings, before its market price shot up, before analysts wrote their commentary of their latest filings, before all of that, there were movements of tankers and drillers and pipeline traffic. They happened months before those activities were reflected in the other data types. What truly characterizes alternative data is that it is primary information, by that we mean the information has not made it to the other sources. 

 There are some problematic aspects of alternative data. It may bring high cost and privacy concerns, in fact it could even bring legal concerns. Capture those data may be very expensive, and the observed company may object, not to mention bystanders. Furthermore, it is usually hard to process. However, Alternative data offers the opportunity to work with unique datasets. Data that is hard to store, manipulate, and operate may be very rewarding, because perhaps your competitors did not try to use it for logistic reasons, gave up midway, or processed it incorrectly. Therefore, you may have a truly unexploited field to mining.

\subsection{Autonomous Analysts}
\label{sec:analyt}

Investment analysis can be seen as the process of assessment and selection of investments options in terms of risk and return in order to provide advice to the investor or manager. Analysts are responsible for transforming raw data into information that can guide investment decisions. The small but relevant difference from traders is that analyst are independent from investor preferences, investment policies or the amount of capital to be managed. Analysts are dependent of the data related to their target assets. It is important to observe that some data may be very relevant for some asset, while it is irrelevant or almost to another asset. For instance, oil prices can be very relevant to airline companies, but perhaps they are not that relevant to bank stocks.  Trading analysis for cryptocurrency prices are most based on historical prices and volumes or some times in signal obtained from news using natural language processing methods (sentiment analysis). In comprehensive survey about cryptocurrency with 126 research papers, no one used oil price to predict cryptocurrency prices~\cite{Ventre:20}. We do not argue that using oil prices would be a good idea, our point is that autonomous analysts for different assets may require different kind of information besides historical prices and volume.

Meanwhile, autonomous trading is the process of submitting buy or sell orders by an autonomous software agent software in order to achieve financial goals. It is quite common that autonomous trading agent deals with just one asset~\cite{Castro:13}. However, it is well know in finance field that an investor must consider the relationship among the investments to build a portfolio that will best meet  the investor's objectives~\cite{Brown:12}. It seems that developers of autonomous traders often disregard two important aspects: investor preferences and the relationship among assets. We address these issues in section~\ref{sec:apm} and we use the term \textbf{autonomous portfolio managers} (APM) to differentiate autonomous agents that deal with such issues from those who do not, what we call simply autonomous traders. Some authors do not make such distinction. We discuss increasing order of complexity: autonomous analysts (~\ref{sec:analyt}), autonomous traders (section~\ref{sec:atrader}) and autonomous portfolio managers (section~\ref{sec:apm}).

\subsubsection{Overfitting}
\label{sec:overfitting}

The problem of creating models with great performance in some known data set, but with bad performance in new data is well known in Machine Learning field and it is usually called \textbf{overfitting}. Nevertheless, this problem seems to be unknown or at least disregarded by many practitioners in autonomous analysis and trading. However, the dangers of failing to avoid overfitting are even more severe in the financial field for two reasons. Overfitting is more likely to happen in finance than in traditional machine learning problems, like face recognition, because of the low ratio signal-noise and the fact that markets are not IID (independent and identically distributed). Furthermore, overfitting in finance leads to over optimistic expectations about performance~\cite{Prado:14}. For a deeper discussion about the problem, we suggest~\cite{Prado:18,Castro:20}

\subsection{Autonomous Traders}
\label{sec:atrader}

Autonomous traders need to deal with two main concerns: price prediction and capital allocation, which is also called bet sizing~\cite{Prado:18}. More than just achieve a high accuracy, an autonomous trader should increase profits when it is right and reduce loss when it is wrong. Furthermore, price (or trend) prediction is highly dependent of asset's nature and history, while capital allocation is dependent only on investor preferences about risk and return and assets' correlation.

\begin{figure}
	\centering
		\includegraphics[width=0.90\columnwidth]{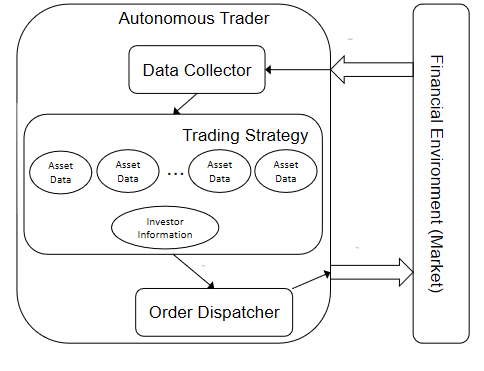}
	\caption{Single Agent Architecture for Autonomous Traders}
		\label{fig:arch1}
\end{figure}

Many studies in autonomous traders (AT) adopt implicitly or explicitly a \textbf{mono agent architecture} that encapsulates most of the complexity in a single component that defines orders given some information, see Figure~\ref{fig:arch1}. It is also common to define one auxiliary module that collects data from external sources that may be relevant to the trading strategy, that we refer as Data Collector. Another module to dispatch the defined orders to the market is also common. Despite the simplicity advantage, such architecture has an important drawback. The trading strategy may become very complex as the number of assets increases, specially if the assets' nature differ very much from each other. Likely it is the main reason that leads such studies to focus on creating AT that deals with just one asset at the time. However, it misses important issues regarding dependence among assets.

As stated before, it is important to reason about dependencies among assets and investor preferences about risk and return. For instance, an investor may be more averse to risk than others, but lower risk portfolio can be obtained by incorporation negatively correlated assets~\cite{Brown:12}. We address autonomous traders that care about these important issues in next section.

\subsection{Autonomous Portfolio Managers}
\label{sec:apm}

The portfolio theory proposed by Markowitz~\cite{Markowitz:52} become the \textit{de facto} standard~\cite{Brown:12}. Under the portfolio theory assumptions, a single asset or portfolio of assets is efficient if no other asset or portfolio of assets offers higher expected return with the same (or lower) risk or lower risk with the same (or higher) expected return. It is possible to achieve lower risk portfolios by including negatively correlated assets in it. In fact, it is possible to derive portfolios that have lower risk than the assets that compose it. This ability to reduce and manage risk is the essence of \textbf{diversification}. Many optimization techniques may be used, such as Linear programming, Convex optimization, Quadratic programming, Meta-heuristic methods, Genetic algorithm and others. Despite all the work done, portfolio optimization is still an open problem and the available methods have some drawbacks. For instance, it is well known that if the mean return vector and the return covariance matrix for the target assets are known, then the Markowitz problem has a closed-form solution. However in practice that is never the case, and therefore they are estimated from historical data. According to some authors~\cite{Pollak:12}, it may turn the Markowitz theory impracticable in real portfolio management applications. We suggest~\cite{Prado:18}, chapter 16, for a deeper discussion about that.

However, it is interesting to observe that autonomous trader developers quite often disregard that fact. According to the portfolio theory, any investor would prefer the efficient portfolio, for a given level of risk or return. However, different investor may have different levels of maximum acceptable risk or minimum acceptable return, therefore they would prefer different portfolios. Autonomous traders should explore diversification and be aware of investor preferences, in order to do a better in trading in the investor's behalf. As stated before, We use the term \textbf{autonomous portfolio manager} to refer to an autonomous trader that uses these facts in its decision process, despite the fact that other authors use the term, autonomous trader, for both cases. 

Autonomous portfolio managers should also reason about \textbf{investor preferences}. Naturally, it include preferences about risk and return, but we have no doubt that ethical, environmental and social concerns will have to be incorporated in autonomous traders in the near future, just like investors are demanding these concerns from human investment managers~\cite{OLeary:21}.

Investor preferences about return and risk may be expressed as a maximum acceptable risk or minimum acceptable return~\cite{Castro:13}, but it could also be expressed in terms of a list of possible assets and percentage limits for some classes of assets. For instance, Graham's book 'Intelligent Investor' suggests an investor should never have less than 25\% or more than 75\% of her portfolio in common stocks and, at the same time, never more than 75\% or less than 25\% in bonds~\cite{Graham:06}. In terms of autonomous portfolio managers, an \textbf{investment policy statement} (IPS) may be defined as a class that implements a method to check if the current portfolio is adherent to IPS and other to check a set of intended orders would make the new portfolio violate the IPS.

When an APM deals with many assets, it is not hard to realize that each asset will demand different data streams in order to be analyzed and it may require a significant aunt of data volume for analysing all assets. One should also note that investor information is also part of data needs to be considered to define the trading strategy. That may bring the so called \textbf{curse of dimensionality}. As the number of dimensions because the amount of data needed to support result  (in a significant way) will likely grow exponentially with dimensionality. In another hand, it is well known that diversification (a large number of target assets) may provide a better risk-reward relation in portfolios, by reducing the risks. So, APMs are required to deal with a large number of assets.

The challenges discussed here: Curse of dimensionality, diversification and investor preferences awareness make the task of creating effective autonomous portfolio managers very hard. It may be even harder if one decides to use the single agent architecture. For these reasons, we propose an \textbf{multiagent architecture}, which is described in section~\ref{sec:arch2}.

\subsection{Multi-agent Architecture for Autonomous Portfolio Managers}
\label{sec:arch2}

There are many challenges for real autonomous portfolio managers (APM). Besides the challenges discussed in section~\ref{sec:apm}: risk mitigation by diversification, dealing with the data high dimensionality and investor preferences awareness, an APM should be able to explain its decisions, just like a human portfolio managers are used to. In fact, the ability to provide explanations for its decision, \textbf{explainability}, is considered to be essential to create artificial intelligence applications that are trusted by users~\cite{Gunning:19}. The interest in creating AI models that can explain their decisions is not new, but it has revived in the last years with machine learning models that are hard to interpret~\cite{Preece:18}. Some authors argue the use of inherently interpretable models rather than searching for methods that can interpret the so called black box models~\cite{Rudin:19}. Considering these challenges, we proposed a \textbf{multiagent architecture} for autonomous portfolio managers that divides these challenges among some autonomous agents that working together make the tasks expected from an APM. Our proposed architecture aims to be inherently interpretable by providing analyses for each asset, configurable investment policies and optimization algorithm that try to achieve the best portfolio for a given objective function. It facilitates reusing well known approaches to create analysts ~\cite{Gu:18} and optimize portfolios, as discussed in section~\ref{sec:apm}. At same time, it reduces the problem of high dimensionality, because each analyst deals only with data related to its asset independently from other analysts or investor preferences.

\begin{figure}
	\centering
		\includegraphics[width=0.90\columnwidth]{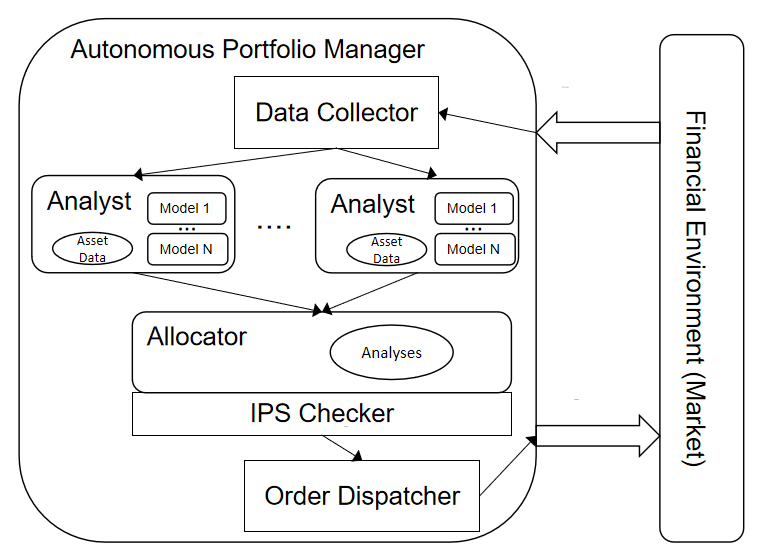}
	\caption{Multi-agent Architecture for Autonomous Portfolio Manager}
		\label{fig:arch2}
\end{figure}

The APM multiagent architecture is presented in figure~\ref{fig:arch2}. It is composed by analyst agents and one allocator agent. An analyst provides a return distribution (analysis) for a given asset using several concurrent models. Using such analyses, the allocator may calculate target prices, expected returns and  optimize the portfolio to maximize an objective function, the Sharpe ratio, for instance. The APM architecture also counts with some additional software components that are based in traditional (non AI based) algorithms. These are shown as rectangles in figure~\ref{fig:arch2}, while autonomous agents are shown as rounded rectangles. These components are described next. 

\begin{itemize}
	\item \textbf{Data Collector}: it collects relevant data about all target assets that may be used by the analysts and their models. It parses and integrates information from different data sources in a format that can be used by the analysts.
	\item \textbf{Analyst}: Autonomous agents that provide an analysis about the asset, ideally a return distribution. Naturally, it is dependent of the asset, but it is independent of investor's preferences and capital. Ensemble several models to create an analyst is a good idea, since it can provide better predictions but also help to understand the reason behind its predictions. The analysts estimate return distributions for each asset using their best effort and available information. 
	\item \textbf{Allocator}: It uses the analyses provided by the analysts and according with investor profile it searches for the best allocation (or re-allocation) of resources among the target assets. It does not need to know more about the assets itself, since it works with the analyses, therefore given the analyses, it is asset independent.
    \item \textbf{Investment Policy Statement (IPS) Checker}: The set of orders are checked to ensure that they respect the investment policy statement. Orders can be ignored or changed by the IPS checker to warrant its goal.
    \item \textbf{Order Dispatcher}: Once the orders coherent with IPS are defined, these orders need to be dispatched to some Market (real or simulated) to be completely or partially executed or even not executed according to the market conditions. That is the role of the \textit{Order Dispatcher}.
 \end{itemize}

The multiagent architecture may be seen as a very simplified structure of a hedge fund, which could be described in three parts: analysts team, portfolio manager and traders. The analysts make analysis of the target assets and pass it to the manager. The manager checks the portfolio composition and makes new positions based on their risk control strategy and the analysts' analysis.  Ultimately the positions updates are executed by traders. Naturally, it is an overly simplified explanation about a hedge fund work process.

\subsection{Performance Measure for Autonomous Analysts and APMs}
\label{sec:perf}

Autonomous Analysts as described earlier are concerned with asset price prediction. Therefore it is easy to note that their performance may be measured as traditional machine learning models for classification (in case price is discretized) or regression (continuous price). Concepts like accuracy, mean-square error, cross-validation and other common associated to machine learning evaluation are all relevant, but one should be special careful to avoid overfitting since it is more dangerous and likely in finance~\cite{Prado:18}. Furthermore, one should consider taking into account the cost of errors. If an Autonomous Analyst (AA) predicts a high return and the return is even higher than expected, it may be considered an error, but certainly if the real return were negative it would be a much worse error. For a deeper discussion about cost of errors in autonomous trading see~\cite{Castro:18}. On the other hand, in order to address evaluation of autonomous traders or APM, we need to review evaluation of human portfolio managers.

There are many possible procedures to evaluate portfolio manager's performance beyond the simple historical return comparison with other managers. Such comparison may be useful, but it does not provide a comprehensive evaluation of the manager's performance. It is absolutely necessary to control for the portfolio risk~\cite{Brown:12}. However, it is quite common to observe in evaluation of autonomous trader without a proper control for the risk of the portfolio. We argue that there is no good reason to evaluate autonomous traders (or autonomous portfolio managers) using different methods than those used for human professionals. Despite the fact that there is not a single universally accepted method for evaluating portfolio performance adjusted to risk, there are several techniques that are used in practice, among them Sharpe's ratio~\cite{Sharpe:94}. It is widely used despite some criticism about it. For instance, high outlier returns could increase the value of the Sharpe ratio's denominator (standard deviation) more than the value of the numerator (return in excess), that way the ratio would be lower thereby lowering the value of the ratio and for positively skewed return~\cite{Rollinger:13}. Some alternative methods are also available, as for instance: Sortino ratio~\cite{Sortino:96} and Modigliani index~\cite{Damodaran:10}. Furthermore, it is important to measure performance for a proper track record length~\cite{Bailey:12}. There is a solid literature about evaluation of portfolio performance~\cite{Brown:12}, but it goes beyond the scope of this paper to cover it here.

\section{Comparative Analysis of Frameworks for Autonomous Traders}
In table~\ref{tab:comp}, we present a comparative analysis of some selected systems with similar propose of mt5se. Such analysis is based on some features that facilitate the development and test of autonomous trading strategies. We do not intend
to judge the overall quality of the cited systems, but just identify differences (positive
and negative) with the mt5se framework proposed here. 
It is relevant to note that mt5se relies on MT5~\cite{Metatrader:18} to connect to stock markets. The MT4~\cite{mt4:20} is not just an earlier version of MetaTrader 5, but an alternative tool focused on Forex markets. While, MetaTrader 5 allows trading in Forex or stock markets, however it comes with the cost of higher complexity and two accounting systems: netting and hedging systems. Therefore, MT4 is still wildly used for Forex operations. The other system, called AgEx~\cite{Castro:09}, is an open source financial market simulation tool that uses FIPA ACL language and a market ontology created specifically to be used for trader agents.

\begin{table}[ht]
	\begin{center}
		\begin{tabular}{|| l | l | l | l |  l || }  
\hline	 Feature  &	MT4 & AgEx & MT5 &  mt5se \\
\hline  Real Operation support & Yes & No & Yes & Yes     \\
\hline  Stock operation support & No & Yes & Yes & Yes \\
\hline  Forex operation support & Yes & No & Yes & Yes \\
\hline  Cryptocurrency operation support & Yes & No & Yes & Yes \\
\hline  Open source & No & Yes & No & Yes    \\
\hline  Python support & No & No & No & Yes    \\
\hline  Object Oriented & No & Yes  & Yes   & Yes \\
\hline  Backtest support & Yes & Yes & Yes &  Yes \\
\hline
		\end{tabular}
	\end{center}
	\caption{Comparison among Selected Systems }
	\label{tab:comp}
\end{table}

\section{Building Autonomous Traders using mt5se}
\label{sec:mt5se}

In this section, we present how to build and test autonomous traders using the mt5se framework. It allows access to price data (open, close, high, low) and book data (bid, ask) and order placement in simulated or real financial markets. It was designed to operate in any financial market accessible through Metatrader platform, that uses the so called netting accounting system. It was tested in simulated and real operation connected just to the Brazilian stock exchange (B3) and Nasdaq Exchange, but in principle, it could used to trade in any stock exchange with Brokers that provide access to MetaTrader platform. There are several Brokerage company that provide such access for Nasdaq for instance, TradeView~\cite{tvmarkets:21} and in B3, XP~\cite{xp:21}, Clear~\cite{clear:21}, among others.

The framework mt5se may be used to create simple trading robots based on algorithms, but it also supports the creation of traders based on Artificial Intelligence techniques, such as decision trees~\cite{Santos:18}, neural networks~\cite{Tan:08}, Bayesian dynamic networks~\cite{Kolm:14}, reinforcement learning~\cite{kearns:06} and so on. 

In listing~\ref{lst:random}, we present a \textbf{very simple} autonomous trader built using mt5se. It is able to trade with any number of target assets, but it does that randomly. It select aleatory a number of shares between and 1 and 1000, and then also aleatory picks buying or selling that number of shares at market price for each asset, at each moment.

\begin{lstlisting}[language=Python, label=lst:random,  caption=Random Autonomous Trader Example]
import numpy.random as rand
class RandomTrader(se.Trader):
    def trade(self,dbars):
        orders=[] 
        assets=list(dbars.keys())
        for asset in assets:
            if rand.randint(2)==1:     
                order=se.buyOrder(asset,100)
            else:
                order=se.sellOrder(asset,100)
            orders.append(order)
        return orders
\end{lstlisting}

\subsection{Installation and Further information about mt5se}

The framework mt5se may be download from github or PyPI platforms. In its repository~\cite{mt5se:21}, there are also available tutorials presenting mt5se API and how to create simple autonomous Trader, and also Traders based on AI techniques. Some jupyter notebooks with several examples are also available. In order to install mt5se in a Python, you may just use one of the ways presented in listing~\ref{lst:install}

\begin{lstlisting}[language=Python, label=lst:install, caption=Installing mt5se]
# this package is required by mt5se
pip install MetaTrader5  
# installing mt5se package
pip install mt5se

# within a jupyter notebook, you may use:
#import sys
#!{sys.executable} -m pip install Metatrader5  
#!{sys.executable} -m pip install mt5se  
\end{lstlisting}

\subsection{Testing Autonomous Traders}

Once you have built an autonomous trader, you need to verify if it is suitable for operation. The basic form of testing an autonomous trader is often called backtest. It is a kind of evaluation for trading robots. It is basically a trading robot executing with  historical  price  series , and its performance is computed.  
In backtesting, time is discretized according with bars and mt5se controls the information access to the Trader according with the simulated time. As simulation time advances, the function 'trade' is called and receives the new bar info and decides which orders to send. In order to backtest one strategy, one just need to create a Trader, establish the test parameters and execute it. These parameters define trading guidelines (target assets, available capital and horizon of investment) and some system parameters that define operation and log register. In listing~\ref{lst:bckt}, we present an example of backtest definition and execution.

\begin{lstlisting}[language=Python, label=lst:bckt, caption=Backtesting an Autonomous Trader]
# trading data options 
capital=100000
results_file='data_equity_file.csv'
assets=['PETR4','VALE3','ITUB4']

#backtest options
prestart=se.date(2019,12,10)
start=se.date(2019,1,10)
end=se.date(2019,2,27)
# Use True if you want debug information for your Trader 
verbose=False 
#sets the backtest setup
period=se.DAILY 
 # it may be se.INTRADAY (one minute interval)
 
bts=se.backtest.set(assets,prestart,start,end,period,capital,results_file,verbose)
#create trader instance
trader=RandomTrader()
# Running the backtest
df= se.backtest.run(trader,bts)   

\end{lstlisting}

\begin{figure}[htbp]
	\centering
		\includegraphics[width=0.50\columnwidth]{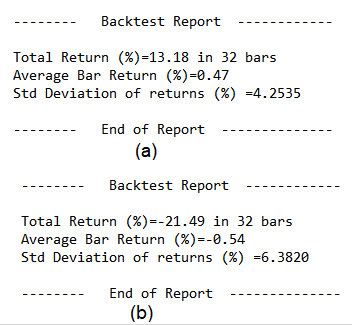}
	\caption{Two different results for the same Trader and setup}
		\label{fig:result_random}
\end{figure}

If we execute the listing~\ref{lst:bckt}, we backtest the trader presented in listing~\ref{lst:random}. As we said before, the random trader does make intelligent decisions, just picks randomly numbers of shares and sells or buys them. Naturally, its performance may vary widely. In order to evaluate the trader's performance, you may use the function se.backtest.evaluate that generates a report, see listing~\ref{lst:eval}. In figure~\ref{fig:result_random}, we present two very different results for the Random Trader using the same assets and trading period. In execution (a), the performance is very good, where trader achieves profit (13.18\% return) in less than two months (January 10 to February 27, 2019), however another execution (b) using the same random trader and period achieved a negative return of 21.49\%.

\begin{lstlisting}[language=Python, label=lst:eval, caption=Evaluating performance an Autonomous Trader]

#Run the trader according setup and get the results
df=se.operations.run(trader,ops)
#evaluate the results
se.backtest.evaluate(df)
#Alternatively, you can evaluate using the generated file
#se.backtest.evaluateFile(fileName)  
#fileName is the name of file generated by the backtest

\end{lstlisting}

We need to note that it is hard to perform meaningful evaluations using backtest. There are many pitfalls to avoid and it may be easier to get trading robots with great performance in backtest, but that perform really badly in real operation. For a deeper discussion about trading strategies evaluation, we suggest~\cite{Castro:20},~\cite{Prado:18} and~\cite{Bailey:12}.

After achieving good performance in backtesting properly conceived and executed, the next phase would be to operate in real market with limited portfolio. As mentioned before you can use mt5se traders in real environments, but you may have to perform some minor changes in a trader so it can operate in real mode and you need to define the operational parameters, which are a little different from backtest parameters. In listing~\ref{lst:ops}, we show how to establish the parameters for trader's operation, create a trader instance from a class called MultiAssetTrader and run it according with its setup. Some brokers provide dedicated servers and accounts for simulated operations, often called demo accounts. For instance, XP Inc is one of those brokers in B3 stock exchange that provide demo accounts. Despite being simulated account, under the point of view of an autonomous trader they are just like real accounts and can use exactly the same code, as presented in listings~\ref{lst:ops} and~\ref{lst:real_trader}. 

\begin{lstlisting}[language=Python, label=lst:real_trader, caption=Autonomous Trader suitable for real operation]
class MultiAssetTrader(se.Trader):
    def trade(self,dbars):
        assets=dbars.keys()
        orders=[]
        for asset in assets:
            bars=dbars[asset]
            curr_shares=se.get_shares(asset)
            money=se.get_balance()/len(assets) # divide o saldo em dinheiro igualmente entre os ativos
            # number of shares that you can buy of asset 
            price=se.get_last(bars)
            free_shares=se.get_affor_shares(asset,price,money)
            rsi=se.tech.rsi(bars)
            if rsi>=70 and free_shares>0: 
                order=se.buyOrder(asset,free_shares)
            elif  rsi<70 and curr_shares>0:
                order=se.sellOrder(asset,curr_shares)
            else:
                order=None
            if order!=None:
                orders.append(order)
        return orders    
\end{lstlisting}

In listing~\ref{lst:real_trader}, we present the complete code of an autonomous trader called MultiAssetTrader, ready for real operations. It is based on a simple interpretation of the Relative Strength Index (RSI), and splits equally the available capital among assets.

\begin{lstlisting}[language=Python, label=lst:ops, caption=Setup and Running an Autonomous Trader for real operation]
      #trading data
# target assets (three of the main assets in B3)
assets=['PETR4','VALE3','ITUB4']
# available capital
capital=100000


# Options for ending time of operation
# endTime=se.now(minOffset=1)   # it will run for one minute!
# endTime=se.now(hourOffset=1,minOffset=30) # the trader will run for 1:30h after started!
endTime=se.operations.sessionEnd() # it will run by the end of session!
  
#if market not open, keep waiting
waitForOpen=True
#get information and decise every minute
timeFrame=se.INTRADAY

#Connect to B3 using default account in MT5
se.connect() 


# System information
# gives information during executing
verbose=True
# operations register file
data_file='data_equity_file.csv'
# seconds to wait between trade calls
delay=1 
# number of bars to take in each decision
mem=10  
# setup operation (ops)  
ops=se.operations.set(assets,capital,\ 
endTime, mem,timeFrame,data_file,\ 
verbose,delay,waitForOpen)

# Create an instance of the trader
trader=MultiAssetTrader()


#Run the trader according setup
 se.operations.run(trader,ops)
\end{lstlisting}

\subsection{Autonomous traders based on AI}
\label{sec:AITrader}
In this section, we are going to present some AI powered trading robots, based on Machine learning and Artificial Intelligence algorithms. In listing~\ref{lst:rand_for}, we present an example based on Random Forests that encompass the whole decision process, taking into account data about target assets and investor information. It is based on Random Forest classifier. Another option would be split those concerns in different models or agents, as discussed in~\ref{sec:arch2}. We present an autonomous trader based on the first alternative, using Random Forests~\cite{Witten:16} and deals with investor preferences in a very simplistic way, by dividing the capital equally among the assets and using the same model in listing~\ref{lst:rand_for}

\begin{lstlisting}[language=Python, label=lst:rand_for, caption=Example of Autonomous Trader based on AI (Random Forest)]
## Defines the Simple AI Trader
from sklearn.ensemble import RandomForestClassifier
from sklearn.preprocessing import KBinsDiscretizer

class RandomForestTrader(se.Trader):

    def setup(self,dbars):
        assets=list(dbars.keys())
        if len(assets)!=1:
            print('Error, this trader is supposed to deal with just one asset')
            return None
        bars=dbars[assets[0]]
        # Data preparation
        timeFrame=10
        horizon=1 # it projects the closing price for next bar
        attr_list=['open','close','MA']

        #getting bars info
        bars=se.get_bars('PETR4',timeFrame*5)
        # creating a new feature
        bars['MA']=se.tech.ma(bars['close'])

        # you may use get_XY providing all info 
        target='close'
        X,y=se.ai_utils.get_XY(bars,attr_list,target,timeFrame,horizon)
       
        discretizer = KBinsDiscretizer(n_bins=3, encode='ordinal', strategy='uniform') 
        # creates the discrete target
        dy=discretizer.fit_transform(y)

        #clf = tree.DecisionTreeClassifier()
        clf = RandomForestClassifier(n_estimators=10)
        clf = clf.fit(X, dy)
        self.clf=clf

    def trade(self,dbars):
            assets=dbars.keys()
            orders=[]
            timeFrame=10 
            horizon=1 
            attr_list=['open','close','MA']
            money=se.get_balance()/len(assets) # shares the balance equally among the assets
            for asset in assets:
                bars=dbars[asset]
                curr_shares=se.get_shares(asset)
                price=se.get_last(bars)
                free_shares=se.get_affor_shares(asset,price,money)
                # get new information (bars), transform it in X
                bars=dbars[asset]
                bars['MA']=se.tech.ma(bars['close'])
                X=se.ai_utils.get_X(bars,attr_list,timeFrame,horizon)
                # predict the result, using the latest info
                p=self.clf.predict([X[-1]])
                if p==2:
                    #buy it
                    order=se.buyOrder(asset,free_shares)
                elif p==0:
                    #sell it
                    order=se.sellOrder(asset,curr_shares)
                else:
                    order=None
                if order!=None:
                    orders.append(order)
            return orders    

# creates instance of the Simple AI Trading
trader=RandomForestTrader()  
\end{lstlisting}

\subsection{Building AI-based Autonomous Traders using historical data}

\begin{figure}[htbp]
	\centering
		\includegraphics[width=0.99\columnwidth]{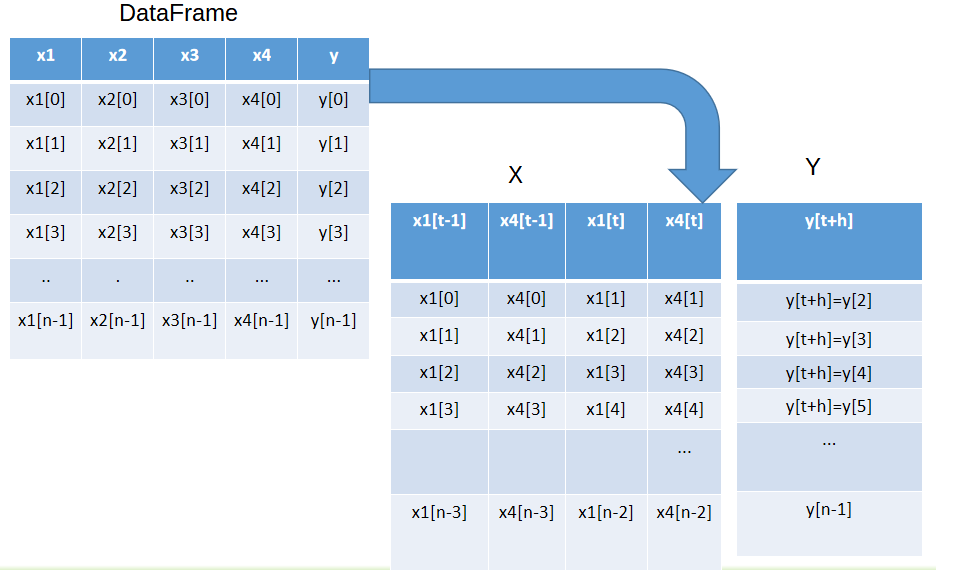}
	\caption{Data Preparation Example: From bars to X and Y features with  time frame (tF)=2, horizon(h)=1 and [x1, x4] selected as relevant features. X and Y are numpy arrays}
		\label{fig:data_trans}
\end{figure}

It is quite common to use historical data to build autonomous analysts or traders using historical data to build Machine Learning models. The training process may take significant amount of computer processing time. The data preparation, which includes gathering, cleaning and selecting features, and also creating new features based on available information, is fundamental and often it requires a lot of effort from practitioners and researches. In this section, we provide some source code examples of Data preparation, ML models creation using mt5se and Sci-kit-learn~\cite{sklearn:12}. In listing~\ref{lst:data_prep}, we present a code snippet that get information about one asset, and creates a supervised dataset with independent features (X) and a target feature (Y), which is the future price of the given asset in a given time horizon. Then, we create an ML model to predict Y given the independent features in a specific time frame in listing~\ref{lst:ml_pred}. The data preparation transform a time series into a suitable dataset to use with sci-kit learn framework, but that could be used in several other frameworks. The figure~\ref{fig:data_trans} illustrates such transformation.

\begin{lstlisting}[language=Python, label=lst:data_prep, caption=Example of Data preparation for Training ML models]
# Data preparation
timeFrame=10
horizon=1 # it projects the closing price for next bar
attr_list=['open','close','MA']

#getting bars info
bars=se.get_bars('PETR4',timeFrame)
# creating a new feature
bars['MA']=se.tech.ma(bars['close'])

# getting X (independent features)
X=se.ai_utils.get_X(bars,attr_list,timeFrame,horizon)

#getting Y (dependent feature)
y=se.ai_utils.get_Y(bars,target,timeFrame,horizon)

# Alternatively, you may use get_XY providing all info 
X,y=se.ai_utils.get_XY(bars,['open','close','MA'],target,timeFrame,horizon)

# Example of Discretization (it may be required by the chosen ML technique
# Discretization

from sklearn.preprocessing import KBinsDiscretizer

discretizer = KBinsDiscretizer(n_bins=3, encode='ordinal', strategy='uniform') 

dy=discretizer.fit_transform(y)  # you make each value assume a discrete value [0,1,2,....] according to discretization strategy

\end{lstlisting}

\begin{lstlisting}[language=Python, label=lst:ml_pred, caption=Creating a ML model for deciding trader's actions and saving it.]
# Creating an AI model
from sklearn.ensemble import RandomForestClassifier
clf = RandomForestClassifier()

# Training an AI based Trader
clf = clf.fit(X, dy)

# Classifing instances
p=clf.predict([X[-3]])
p=int(p)
labels=['sell it','buy it','do nothing']

for i in range(-8,-15,-1):
    p=clf.predict([X[i]])
    l=int(p)
    print('X[',i,']:','decision=',labels[l], ' prediction=',p)
# Now let's save the trained model
import joblib

print(type(clf))
joblib.dump(clf, 'model.joblib')
\end{lstlisting}

\begin{lstlisting}[language=Python, label=lst:ml_model, caption=Creating a ML model for deciding trader's actions and saving it.]
# Creating an AI model
from sklearn.ensemble import RandomForestClassifier
clf = RandomForestClassifier()

# Training an AI based Trader
clf = clf.fit(X, dy)

# Classifing instances
p=clf.predict([X[-3]])
p=int(p)
labels=['sell it','buy it','do nothing']

for i in range(-8,-15,-1):
    p=clf.predict([X[i]])
    l=int(p)
    print('X[',i,']:','decision=',labels[l], ' prediction=',p)
# Now let's save the trained model
import joblib

print(type(clf))
joblib.dump(clf, 'model.joblib')
\end{lstlisting}

\begin{lstlisting}[language=Python, label=lst:ai_pre_trained, caption=Creating a mt5se Trader using a pre-trained Random Forest model]
## Loading a pre-trained model to create an AI based Trader
from sklearn.ensemble import RandomForestClassifier
from sklearn.preprocessing import KBinsDiscretizer
import joblib

class RandomForestTrader(se.Trader):

    def setup(self,dbars):
        fileName='model.joblib'
        self.clf=joblib.load(fileName)

    def trade(self,dbars):
            assets=dbars.keys()
            orders=[]
            timeFrame=10 # it takes into account the last 10 bars
            horizon=1 # it projects the closing price for the next bar
            attr_list=['open','close','MA']
            money=se.get_balance()/len(assets) # shares the balance equally among the assets
            for asset in assets:
                bars=dbars[asset]
                curr_shares=se.get_shares(asset)
                price=se.get_last(bars)
                free_shares=se.get_affor_shares(asset,price,money)
                # get new information (bars), transform it in X
                bars=dbars[asset]
                bars['MA']=se.tech.ma(bars['close'])
                X=se.ai_utils.get_X(bars,attr_list,timeFrame,horizon)
                # predict the result, using the latest info
                p=self.clf.predict([X[-1]])
                if p==2:
                    #buy it
                    order=se.buyOrder(asset,free_shares)
                elif p==0:
                    #sell it
                    order=se.sellOrder(asset,curr_shares)
                else:
                    order=None
                if order!=None:
                    orders.append(order)
            return orders    

# creates instance of the Simple AI Trading
trader=RandomForestTrader()  
\end{lstlisting}

Giving a ML model already trained, the trader example based on Random Forest presented in listingc~\ref{lst:rand_for} can be simplified. It does not train the ML model in the setup function, instead it simply loads the pre-trained model as presented in listing~\ref{lst:ai_pre_trained}, the remaining code stays the same presented  in listingc~\ref{lst:rand_for}.
\section{Examples of Autonomous Traders and Portfolio Managers using mt5se}
\label{sec:results}

In this section, we present the results achieved by implementing three autonomous traders and one Autonomous Portfolio Manager. We have executed some backtest and present the achieved results. In order to build such examples and backtest them. We have implemented a open source python framework, called mt5se. It provides some basic algorithms and allows backtest using historical prices or real time operation. Such operation may be in simulated or real accounts. The framework mt5se uses MetaTrader 5 platform to access real markets. We provide more information about mt5se framework in appendix~\ref{sec:mt5se}. 

\subsection{Backtest setup}
We executed twelve scenarios composed by three strategies (table~\ref{tab:robots}) and four different assets (table~\ref{tab:assets}), over the period of five quarters of trading data (Oct-1-2018 to Dec-31-2019). 

\begin{table} [ht]
	\begin{center}
		\begin{tabular}{|| l | l |l || }
\hline	 \# & ID 	&	 Strategy name 	\\
\hline	1 & RSI & Relative Strength Index	\\
\hline  2 & MA & Moving Average				\\
\hline	3 & RFOR & Random Forest\\
\hline
		\end{tabular}
	\end{center}
	\caption{Selected Trading strategies}
	\label{tab:robots}
\end{table}

\begin{table} [ht]
	\begin{center}
		\begin{tabular}{|| l |l |l  || }
\hline	 \# &	Symbol 	&	 Description		\\
\hline  1 &  AAL & American Airlines Group \\
\hline  2 & AMD & Advanced Micro Devices \\
\hline  3 & MSFT &  Microsoft \\
\hline  4 & UAL	&	 United Airlines Holdings \\
\hline
    	\end{tabular}
	\end{center}
	\caption{Four Selected assets from the most traded assets in Nasdaq }
	\label{tab:assets}
\end{table}

\subsection{Results}

In the next tables, we present the achieved results using the three trading algorithms in each of the four assets in the evaluated period (Oct-1-2018 to Dec-31-2019). For each simulation defined by scenario, stock and trading algorithm, we observed the following values: Annualized Return, Annualized Sharpe Ratio and Volatility. We present the achieved results in tables~\ref{tab:Results1},~\ref{tab:Results2},~\ref{tab:Results3} and~\ref{tab:Results4}.

\FloatBarrier
\begin{table} [ht]
	\begin{center}
		\begin{tabular}{|| l |l | l | l || }
            \hline	 Value - Strategy   & MA      & RSI  &  RFOR \\
\hline   An. Return (\%) & -26.30 & -26.39 & -26.30 \\
\hline   An. Sharpe Ratio (\%) & -18.25 & -18.52 & -18.24  \\
\hline   Volatility (\%)  & 0.822 &0.817 &0.822 \\
\hline
		\end{tabular}
	\end{center}
	\caption{Results for the three strategies for asset AAL}
	\label{tab:Results1}
\end{table}
\FloatBarrier

\begin{table} [ht]
	\begin{center}
		\begin{tabular}{|| l |l | l | l || }
            \hline	 Value - Strategy   & MA      & RSI  &  RFOR \\
\hline   An. Return (\%) & 36.22 & 36.22 &  36.22\\
\hline   An. Sharpe Ratio (\%) & 26.57 & 26.57 & 26.56 \\
\hline   Volatility (\%)  & 1.22  & 1.22 & 1.21 \\
\hline
		\end{tabular}
	\end{center}
	\caption{Results for the three strategies for asset AMD}
	\label{tab:Results2}
\end{table}

\FloatBarrier

\begin{table} [ht]
	\begin{center}
		\begin{tabular}{|| l |l | l | l || }
            \hline	 Value - Strategy   & MA      & RSI  &  RFOR \\
\hline   An. Return (\%) & 28.49 & 28.63 & 28.63 \\
\hline   An. Sharpe Ratio (\%) & 37.51 & 37.48 & 37.48 \\
\hline   Volatility (\%)  & 0.49 &0.49 & 0.49 \\
\hline
		\end{tabular}
	\end{center}
	\caption{Results for the three strategies for asset MSFT}
	\label{tab:Results3}
\end{table}

\FloatBarrier
\begin{table} [ht]
	\begin{center}
		\begin{tabular}{|| l |l | l | l || }
            \hline	 Value - Strategy   & MA      & RSI  &  RFOR \\
\hline   An. Return (\%) & -1.38 & -0.99 & -0.99  \\
\hline   An. Sharpe Ratio (\%) & 3.45 & 3.84 & 3.84  \\
\hline   Volatility (\%)  & 0.63 & 0.62 & 0.62 \\
\hline
		\end{tabular}
	\end{center}
	\caption{Results for the three strategies for asset UAL}
	\label{tab:Results4}
\end{table}
\FloatBarrier

\begin{table} [ht]
	\begin{center}
		\begin{tabular}{|| l |l | l | l || }
            \hline	 Value - Strategy   & MA      & RSI  &  RFOR \\
\hline   An. Return (\%) & 12.80 & 12.82 &  12.85 \\
\hline   An. Sharpe Ratio (\%) & 15.28 & 15.27 &15.27 \\
\hline
		\end{tabular}
	\end{center}
	\caption{Average Results for the three strategies and  four assets}
	\label{tab:media}
\end{table}
\FloatBarrier

\subsection{Autonomous Portfolio Manager Example}
\label{sec:port_opt_impl}
Here, we present an example of an Autonomous Portfolio Manager Example built with mt5se framework. This APM uses three analyst, which source code is presented in listing~\ref{lst:analysts}. Two of them are based on technical indicators (Relative Strength Index - RSI and Moving Average- MA) and the third analyst is based on a Random Forest classifier. These analysts are used as inputs to our APM in listing~\ref{lst:portopttrader} to improve the accuracy of  estimated returns. The APM uses an optimizer to define the capital allocation. This optimizer is based on Sharpe index maximization and uses an implementation provided by the PyPortfolioOpt library~\cite{pyportopt:20}.

\begin{lstlisting}[language=Python, label=lst:analysts, caption=Examples of Autonomous Analysts]
import mt5se as se
import pandas as pd
import numpy as np
from pypfopt import expected_returns
from sklearn.ensemble import RandomForestClassifier
from sklearn.preprocessing import KBinsDiscretizer

class RsiAnalyst(se.Analyst):
    def setup(self,dbars):
        assets=list(dbars.keys())
        df=se.get_close_prices_from_dbars(assets,dbars)
        # train model
        self.mu = expected_returns.mean_historical_return(df)
        self.alpha=0.5
        self.dbars=dbars

    def analyze(self,dbars):
        assets=dbars.keys()
        returns=dict()
        mul=self.mu
        alpha=self.alpha
        for asset in assets:
            bars=dbars[asset]
            # number of shares that you can buy of asset 
            rsi=se.tech.rsi(bars)
            er=self.mu[asset]
            if rsi>=70:  #buy
                exp_ret=er+alpha*abs(er)
            elif  rsi<70: #sell
                exp_ret=er-alpha*abs(er)
            returns[asset]=exp_ret
        return returns   
        
class MAAnalyst(se.Analyst):
    def setup(self,dbars):
        assets=list(dbars.keys())
        df=se.get_close_prices_from_dbars(assets,dbars)
        # train model
        mu = expected_returns.mean_historical_return(df)
        self.alpha=0.5
        self.mu=mu
        self.period=10

    def analyze(self,dbars):
        assets=dbars.keys()
        returns=dict()
        for asset in assets:
            bars=dbars[asset]
            # number of shares that you can buy of asset 
            er=self.mu[asset]
            m=np.mean(bars['close'][-self.period:])
            if se.tech.trend(bars['close'])>0 and bars['close'].iloc[-1]<m:
                exp_ret=er+self.alpha*abs(er)
            elif se.tech.trend(bars['close'])<0 and m<bars['close'].iloc[-1]:
                exp_ret=er-self.alpha*abs(er)
            else:
                exp_ret=None
            returns[asset]=exp_ret
        return returns  

class RandomForestAnalyst(se.Analyst):
    def setup(self,dbars):
        assets=list(dbars.keys())
        df=se.get_close_prices_from_dbars(assets,dbars)
        mu = expected_returns.mean_historical_return(df)
        self.clf=dict()
        for asset in assets:
            bars=dbars[assets[0]]
            # remove irrelevant info
            if 'time' in bars:
                del bars['time']
            timeFrame=10 # it takes into account the last 10 bars
            horizon=1 # it project the closing price for next bar
            target='close' # name of the target column
            ds=se.ai_utils.bars2Dataset(bars,target,timeFrame,horizon)
            X=se.ai_utils.fromDs2NpArrayAllBut(ds,['target'])
            discretizer = KBinsDiscretizer(n_bins=3, encode='ordinal', strategy='uniform') 
            # creates the discrete target
            ds['target']=se.ai_utils.discTarget(discretizer,ds['target'])
            Y=se.ai_utils.fromDs2NpArray(ds,['target'])
            # train model for each asset
            clf = RandomForestClassifier(n_estimators=10)
            clf = clf.fit(X, Y)
            self.clf[asset]=clf
        self.steps=dict()
        self.alpha=0.5
        self.dbars=dbars
        self.mu=mu
    def analyze(self,dbars):
            assets=dbars.keys()
            returns=dict()
            timeFrame=10 # it takes into account the last 10 bars
            horizon=1 # it project the closing price for next bar
            target='close' # name of the target column
            for asset in assets:
                # get new information (bars), transform it in X
                bars=dbars[asset]
                #remove irrelevant info
                if 'time' in bars:
                    del bars['time']
                # convert from bars to dataset
                ds=se.ai_utils.bars2Dataset(bars,target,timeFrame,horizon)
                # Get X fields
                X=se.ai_utils.fromDs2NpArrayAllBut(ds,['target'])
                # predict the result, using the latest info
                p=self.clf[asset].predict([X[-1]])
                er=self.mu[asset]
                if p==2:
                    exp_ret=er+self.alpha*abs(er)      #buy it
                elif p==0:
                    #sell it
                    exp_ret=er-self.alpha*abs(er)
                else:
                    exp_ret=None
                returns[asset]=exp_ret
            return returns    


\end{lstlisting}

\begin{lstlisting}[language=Python, label=lst:portopttrader, caption=Example of Autonomous Portflio Manager using Three Autonomous Analysts (APM)]
import mt5se as se
import pandas as pd
import numpy as np
from pypfopt.efficient_frontier import EfficientFrontier
from pypfopt import risk_models
from pypfopt import expected_returns

class PortOptTrader(se.Trader):
    def setup(self,dbars):
        self.dbars=dbars
        assets=list(dbars.keys())
        df=se.get_close_prices_from_dbars(assets,self.dbars)
        # train model
        mu = expected_returns.mean_historical_return(df)
        S = risk_models.sample_cov(df)
        self.mu=mu
        self.S=S
        #Create analysts
        self.analysts=list()
        anl=se.analysts.RsiAnalyst()
        anl.setup(dbars)
        self.analysts.append(anl)
        anl=se.analysts.MAAnalyst()
        anl.setup(dbars)
        self.analysts.append(anl)
        anl=se.analysts.RandomForestAnalyst()
        anl.setup(dbars)
        self.analysts.append(anl)

    def estimate_exp_returns(self,dbars,mu):
        assets=dbars.keys()
        #execute analysts and update expected returns  
        analysts_mus=list()
        for analyst in self.analysts:
            analysts_mus.append(analyst.analyze(dbars))
        return se.analysts.ensembleAnalyses(analysts_mus,mu)

    def trade(self,bts,dbars):
        order_list=[]
        capital=se.backtest.getBalance(bts)
        assets=list(dbars.keys())
        #for asset in assets:
        #    self.dbars[asset]=self.dbars[asset].append(dbars[asset])
        #    self.dbars[asset]=self.dbars[asset].reset_index(drop=True)
        mu=self.mu
        S=self.S
        mu=self.estimate_exp_returns(dbars,mu)  # get the expected returns from an ensemble of analysts
        ef = EfficientFrontier(mu, S)
        weights = ef.max_sharpe()
        cleaned_weights=ef.clean_weights()
        #create orders using cleaned_weights:
        self.cleaned_weights=cleaned_weights
        # Calculate expected returns and sample covariance
        last_prices=se.get_last_prices(assets,dbars)
        orders=se.volumes_from_weights(assets,cleaned_weights,last_prices,capital,True)
        curr_shares=se.get_curr_shares(assets)
        new_orders=se.get_orders_from_curr_shares(orders,curr_shares)
        for asset in assets:
            if new_orders[asset]>0: #buy it
                order=se.buyOrder(asset,abs(new_orders[asset]),last_prices[asset])
            elif new_orders[asset]<0:  #sell it
                order=se.sellOrder(asset,abs(new_orders[asset]),last_prices[asset])
            else:
                order=None
            if order!=None:
                order_list.append(order)
        return order_list

\end{lstlisting}

In portfolio optimization, it is a common assumption that there is a big amount of capital available and therefore it is possible to diversify your portfolio among many different assets; transactions costs are not relevant and it is possible to invest in a given asset even with a small fraction of your portfolio. These assumptions make sense when you are building a recommendation system for a hedge fund, that manages several billions of dollars. However, that is almost never the case when you are building an autonomous trader. Furthermore, portfolio optimization algorithms often set very small weights for some assets. For these reasons, we propose an algorithm that adjust an optimal theoretical portfolio to a feasible portfolio considering the capital available and the number of units of asset traded in each transaction on an exchange (step). The algorithm returns the orders' volume necessary to adopt the feasible portfolio and it is shown in listing~\ref{lst:order_weight}. The basic idea is buying the closet possible share units to desired weights and redistribute the remain capital, if present, among under allocated assets.

\begin{lstlisting}[language=Python, label=lst:order_weight, caption=From portfolio weights to orders' volume considering capital restriction]
def volumes_from_weights(assets,weights,last_prices,capital):
    # sort weights from highest to lowest
    weights=dict(sorted(weights.items(), key=lambda item:item[1],reverse=True))
    #round 1 - buy while never exceeds the desired weight
    sum=0
    curr=dict() # current weights
    volumes=dict() #  order's volumes
    steps=get_volume_steps(assets)
    for asset in assets:
        aval_capital=weights[asset]*capital
        shares=getAfforShares(asset,aval_capital,last_prices[asset],steps[asset])
        if shares<=0:
            curr[asset]=0
            volumes[asset]=0
        else:
            curr[asset]=(shares*last_prices[asset])/capital
            volumes[asset]=shares
        sum=sum+curr[asset]
    #round 2 - if there is remainig capital, buy more lot according weight order 
    remain_capital=capital-sum*capital
    if remain_capital<=0:
        return volumes
    for asset in weights.keys():
        s=steps[asset]
        p=last_prices[asset]
        missing=(weights[asset]-curr[asset])*capital
        while s*p<remain_capital and s*p<missing:
            s=s+steps[asset]
        if s*p>remain_capital :
            s=s-steps[asset]
        curr[asset]=curr[asset]+(s*p)/capital
        volumes[asset]=volumes[asset]+s
        remain_capital=remain_capital-s*p
        if remain_capital<=0:
            break
    return volumes
\end{lstlisting}

\subsection{Discussion}

As presented in table~\ref{tab:media}, the achieved results by the three strategies  are very similar in terms Annualized return or Annualized Sharpe Ratio. Even when we analyze for individual assets, the performance among strategies is very similar, that happens because when dealing with just one asset it is harder to improve your performance against a competitor that is dealing with the exact same asset. In order to investigate that we implemented two strategies that perform portfolio optimization. The first, called HR, uses the historical returns as prediction for future returns and adjust portfolio weights to maximize its Sharpe ratio using convex optimization. The second, called APM1, uses the previous strategies (MA,RSI and RFOR) as autonomous analysts and ensembles their signals with historical returns to define the expected returns. It is an example of utilization of the alternative multiagent architecture, described in section~\ref{sec:arch2}. When an autonomous analyst signals it is a good time to buy a stock it is interpreted as higher expected return for that stock and a lower expected return when it signals it is a sell moment. Therefore, each autonomous analyst produces an expected return for each asset. The ensemble is performed by simple average.The source code for the Trader with portfolio optimization is available in appendix~\ref{sec:port_opt_impl}. The HR and APM1 traders are just version of the same trader, where HR does not include the three autonomous traders (lines 20-28 in listing~\ref{lst:portopttrader}).

The results for HR and APM1 strategies are presented in table~\ref{tab:ResultsPort}. They used the same period and assets used previously, but it is easy to realize that HR and APM1 performed better than mono-asset strategies. HR and APM1 achieve higher annualized return and Sharpe ratio than any individual strategy. This fact indicates that pursuing multi-asset autonomous strategies may lead to better results. It only confirms that what Markowitz showed years ago about portfolio optimization and diversification is also valid for autonomous traders or autonomous portfolio managers. Nevertheless, it seems that there are significantly more studies in autonomous traders focused on mono-asset traders than multi-asset traders~\cite{Castro:13}.

We can also note that APM1 has performed better than HR. APM1 presented higher Sharpe ratio, higher return and smaller volatility. It indicates that the use of the alternative multiagent architecture (section~\ref{fig:arch2}) that uses specialized analysts to try to find better estimates for future returns, could be an improvement in the efficience of information usage. The framework mt5se helps the development of such autonomous analysts and multi-asset traders.

\begin{table} [ht]
	\begin{center}
		\begin{tabular}{|| l |l | l  || }
            \hline	 Value - Strategy   & HR     & APM1 \\
\hline   An. Return (\%) & 14.84 &  17.06 \\
\hline   An. Sharpe Ratio (\%) & 18.79 & 21.67 \\
\hline   Volatility (\%)  & 0.69 &  0.63\\
\hline
		\end{tabular}
	\end{center}
	\caption{Results for Autonomous Portfolio Management using Historical Returns (HR) and with Autonomous Analysts (APM1) }
	\label{tab:ResultsPort}
\end{table}

We should also note that backtest period (Oct-1-2018 to Dec-31-2019) does not include severe crisis, as the one observed in 2020 or 2008. If that was the case, the performance would be much worse. In fact, we believe there are many problems to be address in autonomous trading in order to build traders that are able to perform as the best human experts, for a deeper discussion, we suggest~\cite{Prado:18}. We address some promising technologies for tackling some of these problems in the next section.

\section{Future of AI in Finance}
\label{sec:probs}

The financial environment is very challenging for autonomous software, however, there are also some promising technologies and some subtle advantages in autonomous traders. In fact, many financial institutions are using autonomous traders as part of the decision process and also trading real portfolios. Nevertheless, there is a long road ahead in the path to build autonomous traders that can beat the best human experts in a consistent way.

One aspect that we believe may be a fundamental advantage for autonomous traders is accountability. Especially regarding eliciting possible conflict of interests. It is well known that there are possible conflicts of interest among analysts, managers and investors. One common conflict of interest may happen among managers and stockholders~\cite{Damodaran:10} (pg.12). The conflicts of interest among analysts and investors, may take place when analysts have investments on target assets themselves or are contracted by securities emitters. In fact, SEC (U.S. Securities and Exchange Commission) has
a long history of examining potential conflicts of interests among such roles, for more information see~\cite{SEC:16} and~\cite{Walter:03}. Due to the fact that machine can have controlled or at least formally verifiable interests through software verification and validation, possible conflict of interests can be avoided or at very least controlled in a more efficient way. On the other hand, the use of autonomous traders require that trust can be established in its development, deployment, and operation. It is a challenge faced by AI in many scenarios and it is not different in autonomous traders. The concept of trustworthy artificial intelligence has five foundational principles according to~\cite{Thiebes:20}: (1) beneficence, (2) non-maleficence, (3) autonomy, (4) justice, and (5) explicability. One may argue that accountability is strongly related to the first two principles. Autonomy refers in large extent on the promotion of human  oversight (e.g., Guidelines), others also consider the restriction of AI-based systems’ autonomy, where humans retaining the right to decide when to decide at any given time. The justice principle is not to be understood judicially, as in adhering to laws and regulations, but instead in an ethical way. For instance, the utilization of AI should to amend past inequities like discrimination of any kind. The last principle, explicability is without any doubt, critical and challenging for autonomous traders, because given the environment complexity mistakes will happen and the ability to explain and justify past decisions is crucial, in order to build and keep trust in the system. 

When reasoning about autonomous investments one may ask What would happen if autonomous investment analysts or managers become ubiquitous. We believe the scenario described by Fama in his Efficient Market Hypothesis (EMH)~\cite{Fama:70} would take place. The EMH states that financial markets are efficient in pricing assets. Asset prices would reflect all information publicly available and the collective beliefs of all investors over the foreseeable future. Thus, it would not be possible to overcome the performance of the market, using information that is known by the market, except by simple chance.

We have discussed the financial environment complexity in section~\ref{sec:modelling}. In fact, it is a challenging environment not just for autonomous agents, but also for human experts. Some argue that it would be beyond the limits of algorithms. As stated by Marks, "Valid approaches work some of the time but not all. And investing can’t be reduced to an algorithm and turned over to a computer. Even the best investors don’t get it right every time."~\cite{Marks:11}. In other words, Marks point to the fact that the environment is not stationary and circumstances rarely repeat exactly. Furthermore, Marks recognizes that psychology plays a major role in financial markets. We believe that true autonomous trader should some how try to model and reason about market psychological aspects. There are some initiatives aiming to walk towards software and hardware that can mimic processes that exist within the human brain, such as intuition and emotional memory concepts~\cite{Crowder:13} and~\cite{Felix:14}. However, there are more questions then answer about how to model and emotions in autonomous agents and certainly it is a topic that requires further research. 

Recent work point to some promising technologies, specially regarding the use of convolutional neural networks and Gated Recurrent Units (GRU)~\cite{Ohara:20}; reinforcement learning~\cite{kearns:06,Pereira:20}, specially when used deep learning architectures~\cite{Kursun:20,Pereira:20b} and ensemble methods~\cite{ALHNAITY:20,Santos:18,Liu:20}. For a comprehensive review of recent work in autonomous analysts, we suggest~\cite{Obthong:20}. There are significant research work in autonomous agents focused on trading with cryptocurrencies, this paper~\cite{Ahmed:21} provides a good review about the theme.

\section{Conclusions and Future Work}
\label{sec:conc}

In this paper, we discussed some fundamental aspects of modelling autonomous traders, their complex environment under the point of view of an autonomous agent. We proposed a multiagent architecture for autonomous traders that care about diversification and investment policies (autonomous portfolio managers). We also discussed the main steps in the developments of systems under such architecture. Furthermore, we presented a framework (mt5se) that helps the development and testing of autonomous traders. This framework mt5se is freely available~\cite{mt5se:21} and it may also be used in real or simulated operation in financial market accessible through platform MetaTrader 5~\cite{Metatrader:18}. We implemented some mono-asset traders and multi-asset traders using Technical indicators and an AI based technique. We performed some backtest and the results indicate that multi-asset traders performed better than mono-asset strategies, which fact indicates that pursuing multi-asset autonomous strategies may lead to better results. It only confirms that what Markowitz showed years ago about portfolio optimization and diversification, is also valid for autonomous traders. Nevertheless, it seems that there are significantly more studies in autonomous traders focused on mono-asset traders than multi-asset traders~\cite{Castro:13}. We also implemented and tested an autonomous trader based on the proposed multiagent architecture that uses the mono-asset traders as autonomous analysts. This trader, called APM1, has performed better than all others. APM1 presented higher Sharpe ratio, higher return and smaller volatility while trading the same assets. Finally, we discussed some open problems in the area such as accountability and trustworthiness in autonomous systems and recognized there is still a long road ahead in the path to build autonomous traders that can beat the best human experts consistently. We also pointed out some interesting technologies that may contribute to advance in such task. The proposed framework mt5se may also contribute to development of new autonomous traders.



\bibliographystyle{unsrt}
\bibliography{bibliopa}

\end{document}